\documentclass{article}

\usepackage[numbers, super]{natbib}
\usepackage{threeparttable}

\makeatletter
\renewcommand\@biblabel[1]{#1.}
\makeatother
\usepackage{float}
\usepackage{graphicx}
\usepackage[utf8]{inputenc} 
\usepackage[T1]{fontenc}    
\usepackage{hyperref}       
\usepackage{url}            
\usepackage{booktabs}       
\usepackage{amsfonts}       
\usepackage{nicefrac}       
\usepackage{microtype}      
\usepackage{xcolor}         

\usepackage[final]{neurips_2023}

\title{StAyaL | Multilingual Style Transfer}

\author{%
  Karishma Thakrar\textsuperscript{*} \\
  Cohere for AI Community \\ 
  \And
  Katrina Lawrence\textsuperscript{*} \\
  Cohere for AI Community \\
  \And
  Kyle Howard \\
  Cohere for AI Community \\
      \And
    \And
      \And
        \And
}

\begin{document}

\maketitle

\renewcommand{\thefootnote}{\fnsymbol{footnote}}
\footnotetext[1]{These authors contributed equally to this work. 
\\ This paper was authored in September 2024 for the Cohere for AI Hackathon (Expedition Aya) and was published on arXiv in January 2025.}
\renewcommand{\thefootnote}{\arabic{footnote}}

\begin{abstract}

  Stylistic text generation plays a vital role in enhancing communication by reflecting the nuances of individual expression. This paper presents a novel approach for generating text in a specific speaker's style across different languages. We show that by leveraging only 100 lines of text, an individuals unique style can be captured as a high-dimensional embedding, which can be used for both text generation and stylistic translation. This methodology breaks down the language barrier by transferring the style of a speaker between languages. The paper is structured into three main phases: augmenting the speaker’s data with stylistically consistent external sources, separating style from content using machine learning and deep learning techniques, and generating an abstract style profile by mean pooling the learned embeddings. The proposed approach is shown to be topic-agnostic, with test accuracy and F1 scores of 74.9\% and 0.75, respectively. The results demonstrate the potential of the style profile for multilingual communication, paving the way for further applications in personalized content generation and cross-linguistic stylistic transfer.

\end{abstract}

\section{Introduction}

The rapid globalization of the modern world has fostered an unprecedented need for effective communication across diverse linguistic and cultural boundaries. Language, as a medium of expression, transcends its functional role as a conduit of information; it serves as a reflection of identity, upbringing, and cultural values. Stylistic text generation, in this context, is an essential tool that enables nuanced communication by preserving the speaker's unique manner of expression. This paper explores generating text in one's style, as well as multilingual style transfer, aiming to bridge the gap between communities by transferring individual styles of speech across languages.

Traditional methods of persona generation and style representation have typically focused on content-based approaches \cite{park2024} \cite{argilla2024finepersonas}. These techniques prioritize capturing explicit information, such as user preferences or values, often relying on extensive input data or self-stereotyping mechanisms. Recent advancements, such as simulation agents and large-scale persona datasets, demonstrate the potential of AI-driven solutions to create rich user representations. However, these methods often overlook the subtleties of speech style, such as tone, rhythm, and linguistic choices, treating it as ancillary to content. This oversight leads to the loss of critical nuances that define an individual's unique expression and fails to capture the richness of a speaker's stylistic identity, particularly when extending such representations to multilingual contexts.

In this study, we propose a novel methodology to create language-agnostic style profiles that capture the intrinsic stylistic features of individuals with minimal input. Our framework leverages modern machine learning techniques to separate the stylistic features of an individual’s speech from the content, using just 100 lines of text to generate a high-dimensional style embedding for each speaker. This embedding is designed to be topic-agnostic and transferable across languages, enabling both stylistically consistent translations and original text generation in the speaker's unique style.

Our approach is grounded in embedding-based representations, clustering algorithms, and supervised contrastive learning. By combining speaker-provided text with carefully selected external datasets and only selecting stylistically consistent data, the speaker data is augmented to create a substantial dataset of stylistically consistent content. This augmented dataset is processed through a Siamese neural network trained on contrastive pairs, with further refinement using a Random Forest Classifier. The resulting style embeddings are evaluated for their capacity to maintain stylistic fidelity while being robust to variations in language and content.

The implications of this research extend beyond individual stylistic representation. Potential applications include personalized communication tools, multilingual conversational agents, and style-driven content generation for diverse contexts. By addressing existing gaps in persona generation and style transfer, this work contributes to the broader goal of enabling richer, more inclusive global communication. The subsequent sections detail our methodology, experimental setup, and results, which collectively underscore the efficacy of our approach in achieving stylistically accurate and multilingual text generation.

\section{Related Work}

\subsection{Existing Approaches}

There is currently extensive work being done on persona engineering. Many institutions are striving to create AI generated personas as representations of individuals. Traditionally one had to self-stereotype in order for the program to generate their persona. However, this led to misrepresentations of individuals as not all aspects of their personality and being were being accurately captured. 

New research from Stanford and Google DeepMind suggests they can create an AI replica of an individual's personality after conducting a 2-hour interview which spans topics like personal history, career, and beliefs. They state that a "two-hour interview is enough to accurately capture [one's] values and preferences" \cite{park2024}. To generate the persona, an agentic approach is taken. The responses from the interviews are used to create simulation agents, which replicate the personality and decision-making tendencies of the individual. The focus of the interviews is primarily on the substance of the responses (the content) rather than the style of speech or manner in which the answers are provided. This approach emphasizes gathering explicit information about beliefs, goals, and experiences \cite{park2024}. To validate the accuracy of the agents, participants and their corresponding agents completed personality tests, social surveys, and logic games. Results showed an 85 percent similarity \cite{park2024}. Simulation agents differ from traditional tool-based agents since focus on mimicking human behaviour and decision-making, rather than being programmed to perform practical tasks.

Other organizations are focused on creating diverse persona datasets. Agrilla has released a dataset called FinePersonas-v0.1 on HuggingFace which contains over 21 million persona definitions/profiles, making it one of the largest publicly available resources for studying user characteristics \cite{argilla2024finepersonas}. The dataset was curated using methods like MinHash deduplication to ensure uniqueness and high-quality representations. Embeddings were generated using Alibaba-NLP's GTE-large-en-v1.5 model, enabling sophisticated clustering and analysis \cite{argilla2024finepersonas}. FinePersonas is useful for tasks like conversational AI, personalized recommendation systems, and social interaction modeling. Users are, however, required to select which persona fits their needs for given tasks. Users can adjust the persona further by fine-tuning on specific text or creating custom prompts that incorporate desired personality traits. This method still requires some degree of self-stereotyping, or selecting a siloed profile. 

In addition to persona engineering, where a person's profile is generated, there is research being done on writing style analysis and identification. A research paper from PAN 2024 explores multi-author writing style analysis, where there goal is to identify all positions where a change in writing style occurs between consecutive paragraphs in a document \cite{pan2024}. The dataset is derived from Reddit posts, and each document consists of paragraphs written by different authors, split into "Easy", "Medium", and "Hard" datasets, with increasing difficulty depending on topic coherence. To identify when a change in author occurred, various approaches focused on semantic and stylistic feature extraction are explored. In particular, RoBERTa is used to encode the text, and contrastive learning for writing style is used, similar to this paper, to separate instances of similar and dissimilar styles \cite{pan2024}. Adaptive entropy-based stability-plasticity techniques were also employed\cite{pan2024}.

Wu et al. proposes a Multi-Channel Self-Attention Network (MCSAN) to explore syntactic and semantic features for authorship attribution \cite{wu2021mcsan}. The study proposes a method that integrates syntactic dependency trees and semantic self-attention mechanisms to analyze writing styles and ultimately identify authors of anonymous documents \cite{wu2021mcsan}. The MCSAN model incorporates a graph neural network for intra-sentence analysis and a self-attention mechanism for inter-sentence relationships, combining these insights for robust authorship classification \cite{wu2021mcsan}. The method outperformed existing state-of-the-art approaches on datasets like CCAT50, demonstrating its effectiveness in distinguishing authors based on their unique writing styles.

\subsection{Gaps in Literature}
Each of the existing works touches on an aspect of either style identification or persona generation. Our paper bridges the gap between the two categories of existing work and adds a multilingual element. There currently is no emphasis placed on extracting style with embeddings with the application of generating text or generating stylistically consistent translations. 

The current persona generation relies on users to self-stereotype or go through a lengthy interview process, which largely analyzes the content of the individual's answers rather than the way the answer is delivered. Persona engineering refers to the systematic process of designing and utilizing detailed profiles or "personas" to represent target user groups\cite{monash2023persona}. These personas are crafted based on data and insights into user behaviours, preferences, and pain points. In practice, persona engineering involves defining user needs, validating personas with stakeholders, and iteratively refining them as user trends evolve\cite{applabx2023userpersonas}.

Persona engineering is not the same as creating an intrinsic style profile. Persona engineering can be thought of as an "outward in" approach, relying first on more generic and external factors, to determine a user profile. Whereas, generating an intrinsic style profile can be thought of as an "inward out" approach, where one's behaviours, activities and preferences are not explicitly defined, but rather extracted through the way in which they speak and what they refer to. Personas are also not curated as translation tools, but as an individual's proxy.

Our paper proposes not only to identify style, but to extract style from an individual's text and use that stylistic representation to generate text in a multilingual manner.  The goal is to capture one's intrinsic style which is a natural construction of their values, upbringing, family, and geographic location. This paper proposes that 100 lines of text is sufficient to achieve the goal, and no self-stereotyping is required. The result will allow users to translate text in their style of speech without nuances being lost, as well as generate text in their unique style, allowing for heightened communication ability.

Embeddings are a fundamental and critical component of the paper. Simply put, embeddings are numerical representations of plain text which can in turn be comprehended and analyzed by Large Language Models (LLMs). They are high-dimensional vectors which encode semantic context and can capture the relationships of data tokens \cite{aisera2024}. Embeddings can be generated at the word level, sentence level, or document level. For this paper, embeddings are generated using Cohere's Embed Multilingual v3 model, which supports over 100 languages. A multi-lingual model is chosen in order to generate language agnostic embeddings, and to capture the style of speech of each speaker across various languages. 

To generate the embeddings, the input text is chunked into sentences and passed into the context window. Cohere's Embed v3 model has a context window of a maximum of 512 tokens. On average, an English word is 1.3 tokens, a French word is 2 tokens, a German word is 2.1 tokens and a Spanish word is 2.1 tokens \cite{Marion2024}. The mandatory hyper parameter “input type” is set to “cluster” as the embeddings are intended to be clustered. This model returns normalized embeddings as a list of floats with a dimensionality of 1024. The returned embeddings can use cosine similarity, Euclidean distance, or dot product as the similarity metric \cite{cohere2023}.

\section{Methodology}
\subsection{Overview}

The goal is to generate a language-agnostic style profile for an individual, represented as a multi-dimensional embedding. This embedding can subsequently be used to generate text in the style of the individual, or generate a translation which transfers the style of the speaker from one language to the other. In order for this to be achieved, the model must first be trained on a large sample size of the speaker's text, and secondly the model must be trained to separate the style from the content of the speaker. 
For ease of use, the speaker will only be required to write 100 lines of text in their unique style. As such the paper will be broken down into three main components. The first component involves augmenting the speaker's data with other data sources that inherently have a similar style to the speaker. The second component requires the use of various deep learning techniques to separate the style from the content of the text. Lastly, the third component requires taking the learned embeddings, and mean pooling them for each speaker to generate a unique style profile in the latent space. 

\subsection{Model Pipeline}
The workflow, depicted in Figure 1, illustrates the sequential process to separate style from content. 

\begin{figure}[htbp]
\centerline{\includegraphics[scale=0.4]{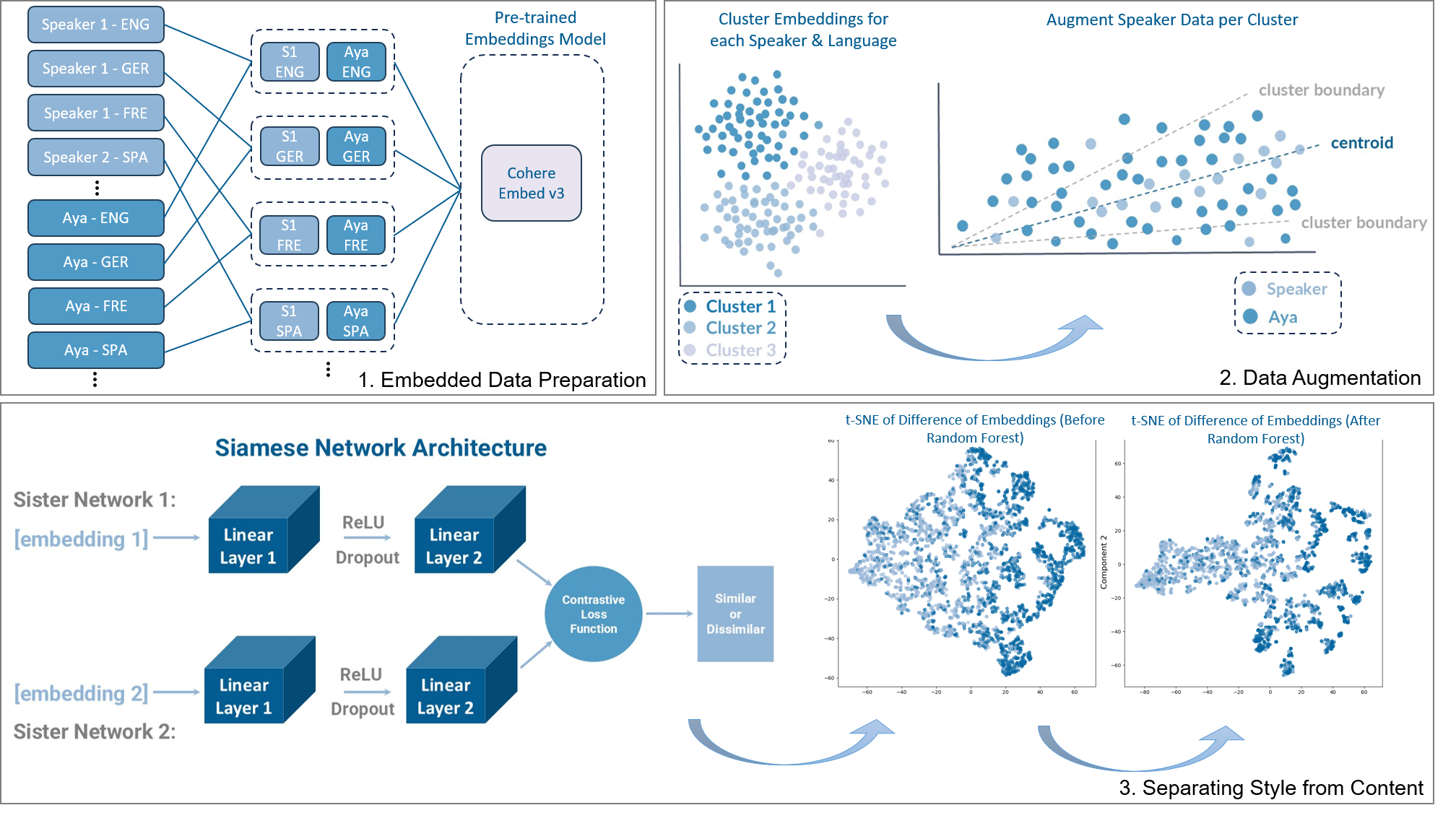}}
\caption{Flow chart of the sequential process of the paper.}
\label{fig}
\end{figure}

The final model, trained on English, French, German, and Spanish, extracts stylistic representations from input text in any language. Using cosine similarity, these representations are compared to a set of generated candidates to identify the text most closely aligned with the speaker’s style. This ensures that the selected output maintains stylistic consistency, enabling text generation that reflects the speaker's unique style across languages.

\subsection{Agglomerative Clustering}

We propose to cluster the embeddings per speaker into semantic groupings such that each cluster represents a distinct topic. Agglomerative clustering is used to achieve this task. Agglomerative clustering is a hierarchical clustering method and is known as a "bottom-up" approach.  As a first step, the algorithm considers each data point (embedding) as its own cluster. To progress, the least distance is calculated between data points, and once determined, a new cluster will form from those data points of least distance.  This process continues as the least distance is then calculated between the centroids of the smaller clusters which will then merge to form new clusters \cite{geron2022}. There are multiple ways distance can be calculated such as cosine distance, Euclidean distance, and Manhattan distance\cite{geron2022}. Additionally, there are four main approaches to determine when the algorithm should halt and stop merging the clusters. 

\begin{enumerate}
    \item Picking a predetermined number of clusters upfront. 
    \item Stop clustering once a cluster of low cohesion is formed.
    \item Stop clustering once a maximum diameter is reached between two points in a cluster.
    \item Stop clustering once the maximum radius of the centroid to a new data point is reached.
\end{enumerate}

We use the fourth approach. Predetermining the number of clusters would impose too great a restriction on the clustering algorithm since the number of topics is not known in advance. Choosing approach number 4 allows to algorithm to operate unbridled.

\subsection{Cosine Similarity}
Cosine similarity, also known as cosine distance, is used to determine the similarity between vectors in multi-dimensional space by calculating the cosine of the angle between them\cite{geron2022}. In particular, it has various applications in Natural Language Processing (NLP) and Machine Learning (ML) as it can be used to determine the semantic similarity between embeddings (how similar the information is between documents or images). Cosine similarity focuses on the direction of the multi-dimensional vectors and not the magnitude, thus it it scale-invariant \cite{geron2022}. It is calculated by computing the dot product of the vectors divided by the product of their magnitudes. The result ranges from -1 to 1, where 1 indicates perfect similarity - cosine of the angle 0 will result in a value of 1 \cite{geron2022}.

\begin{equation}
    \cos(\mathbf{\theta}) = \frac{\mathbf{A} \cdot \mathbf{B}}{||\mathbf{A}|| \cdot ||\mathbf{B}||}
\end{equation}

We applied cosine similarity to identify passages in the Aya dataset that shared stylistic similarities with the speaker text. The purpose of this process is to capture stylistically consistent text in a manner that is context agnostic. Each of the 100 lines of the speaker text is enriched using this method to augment the speaker's style dataset by introducing diverse yet stylistically consistent text. 

\subsection{Contrastive Learning}
Contrastive learning can be performed in a supervised or unsupervised manner. We propose supervised contrastive learning which leverages labeled data to train the models by explicitly defining the similar and dissimilar instances\cite{encord_contrastive_learning}. Contrastive learning operates under the assumption that similar data should be closer in a learned embedding space while dissimilar data should be farther apart\cite{encord_contrastive_learning}. There are multiple loss functions which can be used train the contrastive learning models. Contrastive loss aims to minimize the distance between similar pairs and maximize the distance between dissimilar pairs. This is achieved by mapping similar instances close together in the latent space and pushing apart dissimilar instances\cite{encord_contrastive_learning}. However, it does not directly optimize for classification accuracy. The loss function focuses on the relative distances between pairs rather than the absolute threshold-based classification. If the contrastive loss is decreasing, it means that the model is learning to separate similar and dissimilar pairs in the embedding space. However, decreasing accuracy along side the loss suggests that the separation in the embedding space may not be optimal for classification based on the chosen threshold. The model might be separating the pairs in a way that doesn't align well with the binary classification task.

\subsection{Siamese Neural Networks}
Siamese Neural Networks (SNN) consist of twin networks, each receiving different input data points while sharing the same architecture, weights, and biases. This dynamic weight sharing ensures that the same features are extracted from both inputs, enabling consistent and fair comparisons. Typically, these networks are implemented as deep neural networks, such as Convolutional Neural Networks or fully connected networks.

The inputs to a Siamese Network are provided in pairs, making the architecture particularly suited for contrastive learning, where pairs are categorized as similar or dissimilar. Each network processes its input to produce an embedding, or feature vector, capturing the essential characteristics of the data. A distance metric, such as Euclidean or cosine distance, is then used to compare these feature vectors.

Contrastive loss is a key component of training Siamese Networks. It encourages the network to produce embeddings that are close in the feature space for positive pairs (similar inputs) and far apart for negative pairs (dissimilar inputs). The loss penalizes positive pairs with large distances and negative pairs with small distances, refining the network's ability to differentiate between similarity and dissimilarity in the embedding space\cite{encord_contrastive_learning}.

Below is a general equation for contrastive loss.

\begin{equation}
   L(y, D) = (1 - y) \cdot \frac{1}{2} \cdot D^2 + y \cdot \frac{1}{2} \cdot \max(0, m - D)^2
\end{equation}

In this formula:
\begin{itemize}
    \item \( y \) is the label, where \( y = 0 \) for similar pairs and \( y = 1 \) for dissimilar pairs.
    \item \( D \) represents the distance between the feature vectors of the two inputs.
    \item \( m \) is the margin hyperparameter, defining the minimum distance between feature vectors of dissimilar pairs.
\end{itemize}

During training, the network is fed both positive and negative pairs. The contrastive loss function adjusts the network weights to ensure similar pairs have closer feature representations, while dissimilar pairs are mapped farther apart in the learned embedding space. As a result, the model learns a robust representation where the distances between feature vectors correlate with the semantic similarity of the inputs. The output of a Siamese Neural Network is typically a single value indicating the similarity or dissimilarity between two inputs, making it an ideal architecture for tasks such as verification, matching, and clustering.

In the framework proposed, we employ contrastive learning with Siamese Neural Networks to disentangle style from content in textual data. This approach involves explicitly pairing texts from the same speaker that shared similar stylistic characteristics but differed in content, bringing stylistically similar texts closer together in the embedding space while pushing stylistically dissimilar ones farther apart. By focusing on stylistic similarities, the SNN used in conjunction with contrastive learning allows the embeddings to capture nuanced stylistic identities effectively. This enables a more precise analysis and comparison of style, independent of the content, ensuring that stylistic features are the primary drivers of similarity in the learned embeddings.

\subsection{Random Forest Classifier}
A Random Forest Classifier (RFC) is an ensemble model of Decision Trees. The algorithm works by training a group of Decision Tree classifiers each on a different random subset of the training set, and averaging the predictions of the many trees\cite{hastie2009}. This adds an element of randomness to the model which reduces the risk of over-fitting the data. Individual Decision Trees typically have a high variance, therefore they risk over-fitting the data and they are less robust than an ensemble model\cite{breiman2001}. 

The RFC has all the hyper parameters of a Decision Tree and a Bagging Classifier. The algorithm searches for the best feature among a subset of features, and by default, it does not consider all the features at once\cite{hastie2009}. Random Forests can estimate the importance of features by measuring how much they reduce the impurity in the trees\cite{breiman2001}. As such, the model trades a higher bias for a lower variance, and it is a powerful algorithm that is capable of fitting complex and non-linear datasets. Classification predictions are obtained from all of the individual trees, and the class with the most votes is ultimately taken as the final prediction\cite{geron2022}. 

In this context, we use the RFC to further enhance the stylistic separation achieved by the SNN. By leveraging the embeddings generated through the network, the RFC provides an additional layer of analysis, effectively identifying and classifying patterns based on stylistic differences. This approach allows the embeddings, which encapsulate the stylistic characteristics of the input data, to be validated and refined through the ensemble's classification process. The t-SNE visualizations of the embeddings before and after applying the RFC further illustrate the improved separation of stylistic features, highlighting the classifier’s role in amplifying the stylistic distinctions within the multi-dimensional learned embedding space.

\section{Experiment and Analysis}

\subsection{Datasets}

Data was collected from multiple speakers across their spoken languages, resulting in 15 datasets. To supplement these, data from Cohere for AI's multilingual Aya dataset \cite{singh2024aya} was incorporated, ensuring stylistic relevance. Each of the combined datasets was processed independently to maintain speaker and language-specific consistency.

\subsection{Experiment Setup}

\subsubsection{Data Preparation and Embedding Generation}

We thoroughly cleaned, preprocessed, and reformatted each of the combined 15 datasets. After exploring various embedding models, such a LASER, Cohere’s Embed V3 model, which is language agnostic, is used to embed the text. The sentence chunks are inputted into the context window, and the mandatory hyper-parameter “input type" is set to “cluster” as the embeddings will be clustered. The model output a list of floats with a dimensionality of 1,024. 

\subsubsection{Topic Clustering and Data Augmentation}

Once the embeddings are generated, the intention is to create topic clusters in order to find true style without the noise of semantics. By performing topic clustering, the analysis performed within each cluster will be topic agnostic and will focus on stylistic aspects. Agglomerative clustering, a hierarchical clustering algorithm, is used to group the data into sematic-based clusters. The algorithm will operated unbridled as the number of clusters is not predetermined. 

Evidently, 100 lines of text is not sufficient to train a model to separate the style from the content of the speaker, and not all Aya data will be stylistically similar to the speaker data. As such, only stylistically consistent Aya data will be selected to remain and augment the speaker data. To augment the speaker data, the mean-pooled speaker data in each cluster calculated and set as the centroid. The cosine similarity between the centroid and all embedded text per cluster is the calculated. The minimum of the top 15 or 70 percent of the embeddings with the highest cosine score are kept, and clusters without speaker data are eliminated. Human evaluation is performed for each of the augmented datasets to ensure stylistic accuracy. Please refer to Figure 1 for a simplified visualization of the data augmentation process.

\subsubsection{Language-Specific Considerations}

For the remainder of the paper, we considered only the English, French, German, and Spanish datasets, as the Tamil and Cantonese augmented datasets did not pass the human evaluation step. This outcome reflects inherent language biases and latent gaps in the performance of the language-agnostic embedding model. These gaps are evident in the disparity of clustering results across languages after embedding. For example, embedded French text produced approximately 30\% fewer clusters than English, suggesting reduced granularity in capturing stylistic nuances. Similarly, during data augmentation with cosine similarity, Cantonese data exhibited significantly lower clustering density, with speaker data distributed across only 14 clusters, compared to +70 clusters in English. These discrepancies underscore the limitations of the embedding model in uniformly representing similar stylistic features across different languages.

\subsubsection{Contrastive Learning for Style-Content Separation}

To separate style from content, we employed contrastive learning to train a model capable of distinguishing between stylistically similar and dissimilar pairs. The augmented dataset is pre-processed to form labeled contrastive pairs, defined as follows:

\begin{itemize}
    \item \textbf{Positive (similar) pairs:} same style, different content
    \item \textbf{Negative (dissimilar) pairs:} different style, same content
\end{itemize}

While the ultimate goal is to use contrastive loss with only positive and negative pairs, triplets serve as an intermediary step in the preprocessing pipeline. The first part of the pipeline generates triplets by identifying anchors, positive samples, and negative samples based on speaker and cluster information, ensuring that each anchor has both a corresponding positive and negative example. This approach accommodates cases where suitable positive or negative pairs might not naturally exist. The second step extracts positive and negative pairs from these triplets, ensuring a balanced and well-structured dataset for the contrastive learning task. This method ensures the model is trained effectively while addressing potential data imbalances or gaps during preprocessing.

\subsubsection{Training the Siamese Neural Network}

An SNN (see Figure 1 for architecture) was trained on contrastive learning pairs using contrastive loss to separate stylistic features from content. This approach focuses on maximizing the similarity between positive pairs and minimizing the similarity between negative pairs, enabling the model to isolate common stylistic attributes across languages while de-prioritizing content-specific information\cite{encord_contrastive_learning}. The network effectively organizes the positive and negative pairs in the latent space, with similar pairs pulled closer together and dissimilar pairs pushed further apart.

\subsubsection{Refining with Random Forest Classifier}
To enhance classification performance and separate positive and negative pairs in high-dimensional space, we employed a RFC as a subsequent step. The RFC was chosen for its robustness in handling complex, non-linear decision boundaries, enabling it to operate effectively in the higher-dimensional embedding space learned by the SNN. By leveraging the embeddings produced by the SNN, the RFC is able to further refine the separation of positive and negative pairs, and would result in more distinct clusters with greater accuracy. 

\subsubsection{Creating Style Profiles}

Finally, we developed a style profile per speaker by averaging the learned style representations of text samples from that speaker. This is achieved by having the trained RFC predict the style probabilities for each text sample, distinguishing between positive and negative pairs. The classifier’s learned weights from the final layer are extracted to represent the stylistic attributes of the input text. These representations are then mean-pooled across samples sharing the same style but varying in content, resulting in a comprehensive and robust style profile for the individual.

\subsection{Evaluation Metrics}
The models are evaluated using a combination of human judgment and quantitative metrics to ensure both stylistic accuracy and robust performance. Human evaluation assesses the stylistic consistency and quality of outputs, offering insights beyond numerical metrics. Contrastive loss measures the model's ability to distinguish between stylistically similar and dissimilar pairs, while accuracy captures the overall correctness of predictions. Precision and recall highlight the model's ability to balance true positives and false negatives, with the F1 score providing a comprehensive measure of this balance. Together, these metrics ensure a thorough and meaningful evaluation of the models.

\section{Results}
The SNN, trained on contrastive pairs using contrastive loss, achieves a test loss of 0.1684 and a recall of 0.90. Figure 2 visualizes the loss trajectory across training, validation, and testing datasets over 50 epochs. The training and validation loss curves closely follow each other, indicating effective model generalization with minimal overfitting. The steady decline in loss for both curves suggests the network consistently learns to minimize the distance between similar pairs while maximizing the distance between dissimilar pairs in the embedding space. The SNN demonstrates this effective learning in Figure 3, with clear separation in Euclidean distance between the contrastive pairs. 

To enhance classifier performance, a RFC is employed on the embeddings generated by the SNN. By leveraging the high-dimensional embeddings as input, the RFC capitalizes on the stylistic features learned during contrastive training. As summarized in Table 1, the RFC achieves a test accuracy of 0.7493, with precision, recall, and F1 scores of 0.7512, 0.7545, and 0.7529, respectively. The validation results demonstrate comparable metrics, reinforcing the robustness of the RFC in separating stylistically similar and dissimilar pairs. These outcomes highlight the RFC's ability to effectively classify embeddings in a higher-dimensional hyperplane, enhancing the initial results from the SNN.

\begin{table}[htbp]
    \centering
    \caption{Validation and Test Results for Random Forest Classifier}
    \begin{threeparttable}
        \begin{tabular}{|l|c|}
            \hline
            \textbf{Metric} & \textbf{Value} \\
            \hline
            Validation Accuracy & 0.7531 \\
            Validation Precision & 0.7694 \\
            Validation Recall & 0.7427 \\
            Validation F1 Score & 0.7558 \\
            Test Accuracy & 0.7493 \\
            Test Precision & 0.7512 \\
            Test Recall & 0.7545 \\
            Test F1 Score & 0.7529 \\
            \hline
        \end{tabular}
    \end{threeparttable}
    \label{tab:rf_results}
\end{table}

The separation of positive and negative pairs reflects the model's ability to distinguish style from content, resulting in learned embeddings that serve as a semantic-agnostic representation of the speakers' style. Figure 5 presents a two-dimensional projection of the data on the first two principal components, highlighting the style profiles (mean-pooled speaker embeddings) for each speaker across different languages. While this visualization demonstrates progress, it also reveals areas for improvement. Specifically, embeddings from the same language, irrespective of the speaker, tend to cluster more closely together, suggesting that language-specific biases persist. Addressing these biases and the latent gap in the embedding model will be crucial for enhancing the separation of stylistic representations across languages.

\begin{figure}[htbp]
\centerline{\includegraphics[scale=0.575]{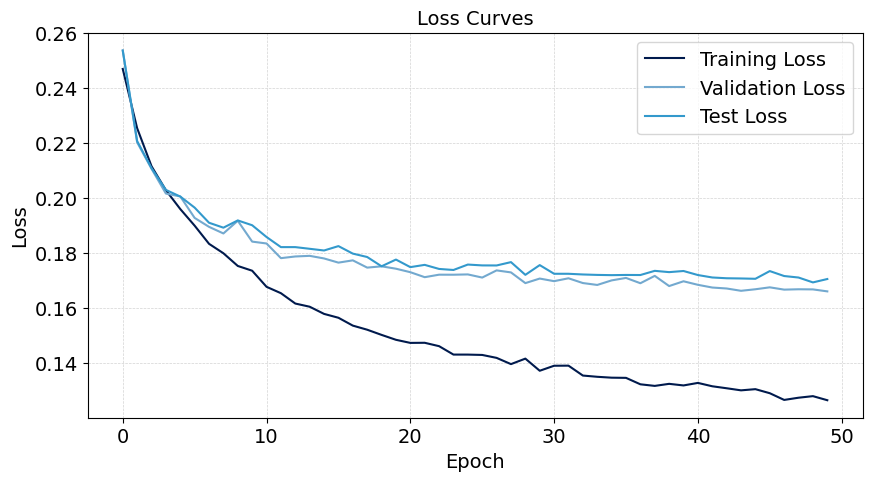}}
\caption{Results of the trained single branch Siamese Network on the contrastive learning pairs. There was a test loss of 0.1684 and a recall of 0.90.}
\label{fig}
\end{figure}

\begin{figure}[htbp]
\centerline{\includegraphics[scale=0.45]{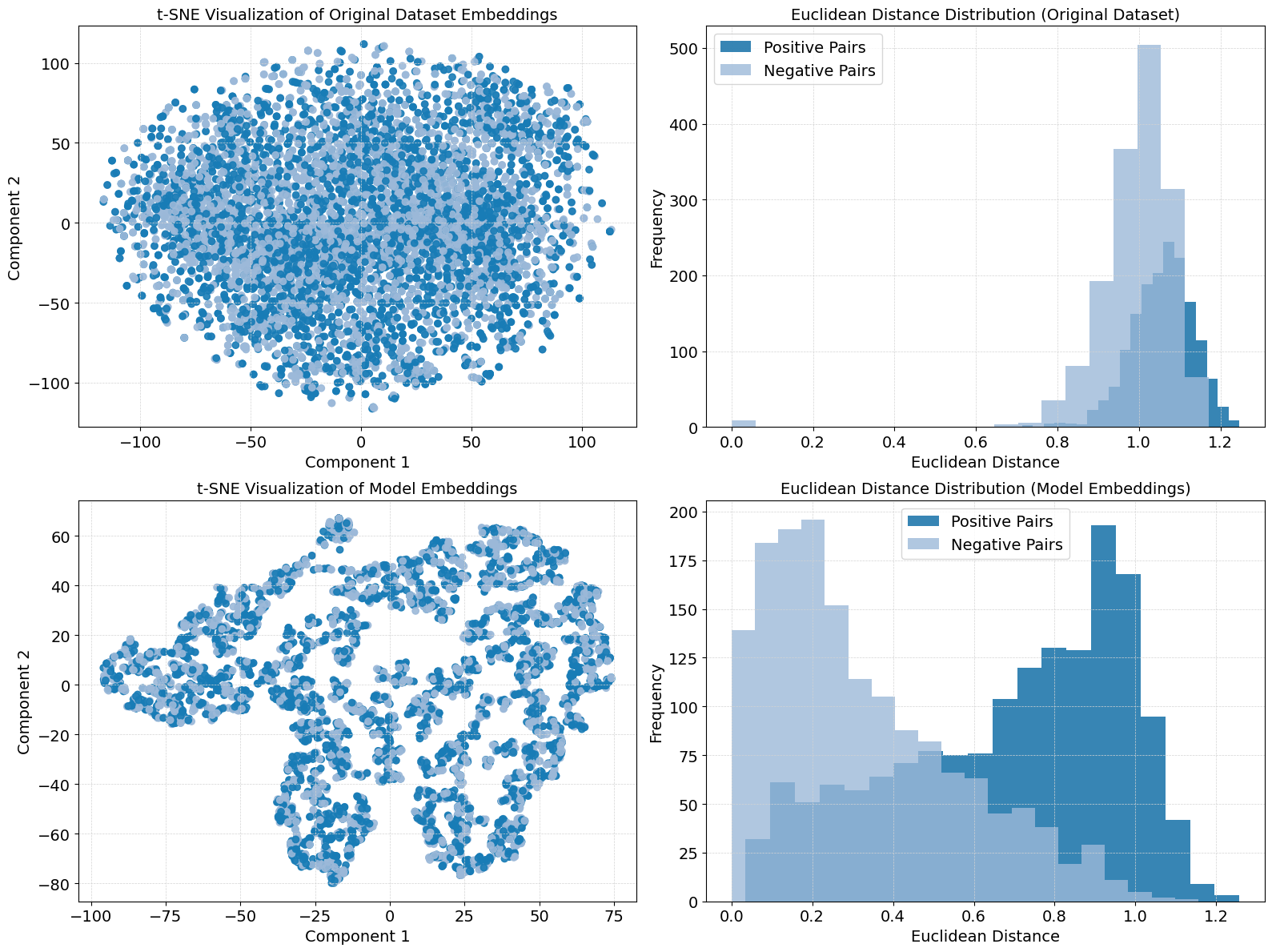}}
\caption{These plots show how the positive and negative pairs are clustered before and after training the Siamese Network on the contrastive pairs. In the learned embedding space, there is a separation between the positive and negative pairs.}
\label{fig}
\end{figure}

\begin{figure}[htbp]
\centerline{\includegraphics[scale=0.40]{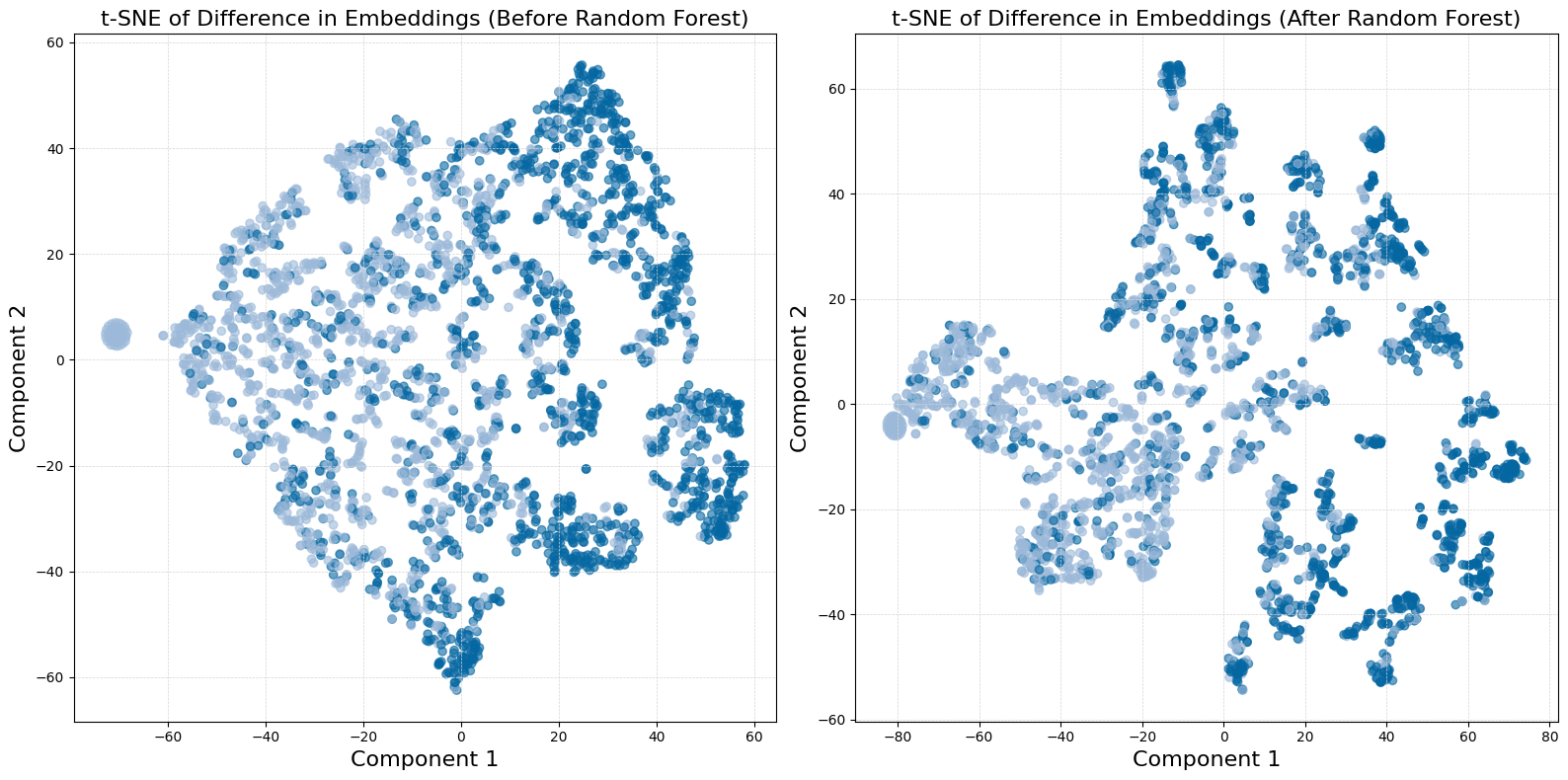}}
\caption{The learned embeddings from the Siamese Network are passed through a Random Forest Classifier to further separate the positive and negative pairs. The plot on the left is a projection of the embeddings before being passed through the classifier. The plot on the right is a projection of the embeddings after they have been passed through the classifier. More distinct groupings (i.e. better separation) can be observed in the second plot.}
\label{fig}
\end{figure}

\begin{figure}[htbp]
\centerline{\includegraphics[scale=0.45]{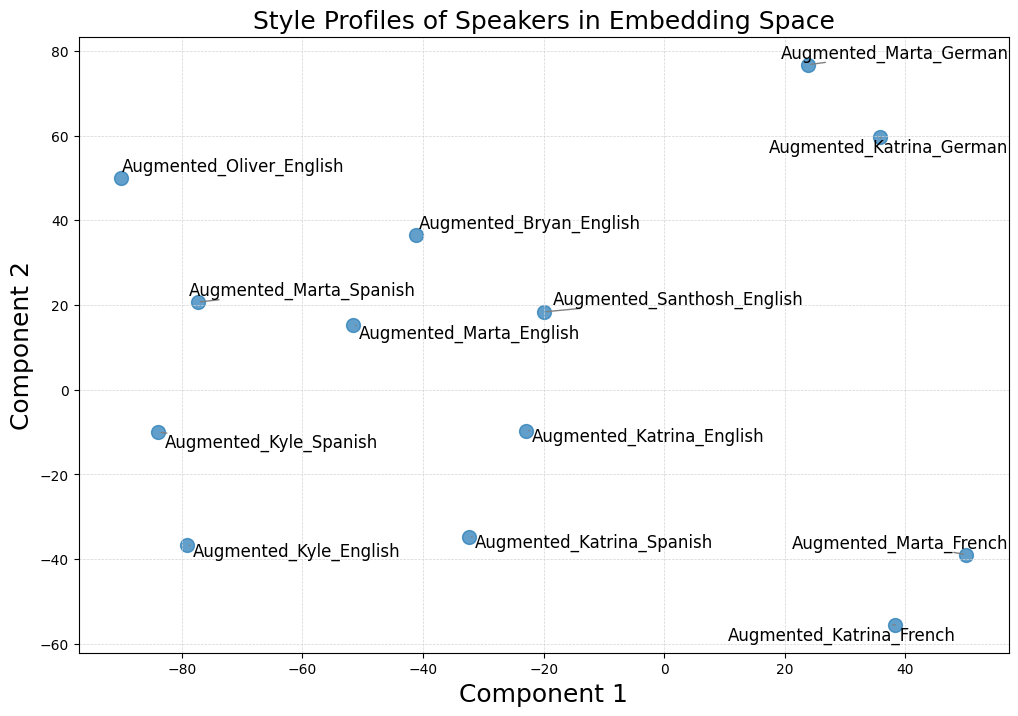}}
\caption{Style profiles of each speaker and in each language in the learned embedding space.}
\label{fig}
\end{figure}

\section{Discussion}
Figure 5 highlights the clustering of style profiles in the latent space, revealing both progress and limitations in the model's representation of stylistic features. Ideally, embeddings for all speakers, regardless of language, would cluster together, indicating successful isolation of style from language. However, the results demonstrate a persistent latent gap. This bias is particularly evident in the closer clustering of embeddings within the same language, which may reflect the model's reliance on linguistic patterns rather than speaker-specific stylistic nuances. Interestingly, the language bias appears less pronounced in certain cases with speakers with a linguistic background, suggesting that individual speaker characteristics may influence the degree of bias observed. The ability of these speakers to maintain consistent stylistic patterns across languages appears to mitigate some of the bias, indicating that individual speaker characteristics can partially counteract language-specific influences. Addressing these biases will be critical for improving the model's ability to produce language-agnostic stylistic representations.

Current limitations in LLMs prevent direct ingestion of embeddings for text generation, requiring alternative approaches to leverage the curated style profiles. Embeddings can effectively determine stylistic similarity between texts but cannot be decoded directly into text. To generate new text that reflects the same style across languages, the model must (a) use the learned language-agnostic style representations to capture stylistic features in new text, regardless of language, and (b) generate multiple outputs, selecting those that best align with the desired style profile.

Future extensions include the potential development of a multi-headed transformer model that integrates both the learned and original embeddings. This approach would preserve critical features from the original embeddings while leveraging the learned style representations, ensuring a more robust and comprehensive style profile. Although the numerical results indicate that the learned embeddings are topic-agnostic, there is concern that some stylistic nuances may be lost during this transformation. Additionally, the RFC, while effective in enhancing separation, may risk over-fitting and memorizing patterns in the training data. Expanding the diversity and size of the original datasets will be crucial to improving robustness and generalization, particularly for out-of-sample instances.

The flexibility of the current framework opens up possibilities for broader applications. Beyond individual speaker profiles, the methodology can be extended to generate collective style profiles for groups. By mean-pooling individual profiles and weighting contributions, group profiles could reflect shared stylistic attributes, such as those of a team, department, or organization. These profiles could enable applications such as capturing and replicating a company’s collective voice, fostering consistent communication aligned with organizational values, or enabling personalized content generation for specific audiences. The versatility of the approach highlights its potential to address both individual and collective stylistic needs across diverse contexts.

\section{Contributions and Team Effort}

This research was a collaborative effort, with significant contributions from all team members. Kyle Howard had the initial idea of stylistic translations. Katrina Lawrence, Karishma Thakrar, and Kyle Howard led the initial ideation phase, focusing on combining and clustering data. Katrina conducted end-to-end research on the mathematical foundations of the algorithms ensure viability and optimal, cohesive performance, including encoding methods, clustering methods, loss functions, neural network architectures and ensemble model architectures. Karishma researched and implemented language-agnostic embeddings for multilingual analysis and researched and built the end-to-end data pipeline and algorithm architecture, including Siamese Neural Network for contrastive learning to separate style from content. Karishma enhanced accuracy by integrating the Siamese embeddings with a Random Forest Classifier and visualized the dimensionality-reduced mean-pooled results of the speaker style profiles. Katrina also developed the cross-validation framework for clustering, ensuring methodological rigor. As team captains, Katrina, Karishma, and Kyle met regularly to align progress. The team's diverse contributions were instrumental in achieving the research goals.

\section{Acknowledgments}

The authors would like to acknowledge Marzieh Fadaee from Cohere for AI for her assistance with editing the paper. The authors would also like to thank Marta Romani, Oliver Bolton, Santhosh G S, and Brian Kwan for providing datasets in their languages which were used in this research.

\newpage
\setcounter{footnote}{1}
\bibliography{mybib} 

\begin{thebibliography}{10}
\expandafter\ifx\csname url\endcsname\relax
  \def\url#1{\texttt{#1}}\fi
\expandafter\ifx\csname urlprefix\endcsname\relax\def\urlprefix{URL }\fi
\providecommand{\bibinfo}[2]{#2}
\providecommand{\eprint}[2][]{\url{#2}}

\bibitem{park2024}
\bibinfo{author}{Park, J.~S.} \emph{et~al.}
\newblock \bibinfo{title}{Generative agent simulations of 1,000 people}.
\newblock \emph{\bibinfo{journal}{arXiv preprint arXiv:2411.10109}}
  (\bibinfo{year}{2024}).
\newblock \urlprefix\url{https://arxiv.org/abs/2411.10109}.

\bibitem{argilla2024finepersonas}
\bibinfo{author}{{Argilla}}.
\newblock \bibinfo{title}{Finepersonas-v0.1 dataset}.
\newblock
  \bibinfo{howpublished}{\url{https://huggingface.co/datasets/argilla/FinePersonas-v0.1}}
  (\bibinfo{year}{2024}).
\newblock \bibinfo{note}{Accessed: 2024-12-06}.

\bibitem{pan2024}
\bibinfo{author}{Ayele, Y.} \emph{et~al.}
\newblock \bibinfo{title}{Multi-author writing style analysis, multilingual
  text detoxification, oppositional thinking analysis, and generative ai
  authorship verification}.
\newblock \emph{\bibinfo{journal}{Proceedings of SIGIR '24}}
  \bibinfo{pages}{1--10} (\bibinfo{year}{2024}).
\newblock
  \urlprefix\url{https://downloads.webis.de/publications/papers/ayele_2024.pdf}.

\bibitem{wu2021mcsan}
\bibinfo{author}{Wu, H.} \emph{et~al.}
\newblock \bibinfo{title}{Exploring syntactic and semantic features for
  authorship attribution}.
\newblock \emph{\bibinfo{journal}{Applied Soft Computing}}
  \textbf{\bibinfo{volume}{107}}, \bibinfo{pages}{107385}
  (\bibinfo{year}{2021}).
\newblock
  \urlprefix\url{https://www.sciencedirect.com/science/article/abs/pii/S1568494621007365}.

\bibitem{monash2023persona}
\bibinfo{author}{University, M.}
\newblock \bibinfo{title}{An introduction to the persona concept and its
  application in practice}.
\newblock \bibinfo{howpublished}{Monash University Seminar}
  (\bibinfo{year}{2023}).
\newblock
  \urlprefix\url{https://www.monash.edu/it/ssc/seminars/2023/an-introduction-to-the-persona-concept-and-its-application-in-practice}.

\bibitem{applabx2023userpersonas}
\bibinfo{author}{Applabx}.
\newblock \bibinfo{title}{What are user personas and how to define them?}
\newblock \bibinfo{howpublished}{Applabx Blog} (\bibinfo{year}{2023}).
\newblock
  \urlprefix\url{https://blog.applabx.com/what-are-user-personas-and-how-to-define-them/}.

\bibitem{aisera2024}
\bibinfo{author}{Aisera}.
\newblock \bibinfo{title}{Llm embeddings}.
\newblock \emph{\bibinfo{journal}{Aisera Blog}}  (\bibinfo{year}{2024}).
\newblock \urlprefix\url{https://aisera.com/blog/llm-embeddings/}.
\newblock \bibinfo{note}{Accessed: 2024-12-12}.

\bibitem{Marion2024}
\bibinfo{author}{Marion, S.}
\newblock \bibinfo{title}{What are openai gpt tokens?}
\newblock \emph{\bibinfo{journal}{GPT for Work}}  (\bibinfo{year}{2024}).
\newblock \urlprefix\url{https://gptforwork.com/guides/openai-gpt3-tokens}.
\newblock \bibinfo{note}{Accessed: 2024-12-12}.

\bibitem{cohere2023}
\bibinfo{author}{Reimers, N.} \emph{et~al.}
\newblock \bibinfo{title}{Introducing embed v3}.
\newblock \emph{\bibinfo{journal}{Cohere Blog}}  (\bibinfo{year}{2023}).
\newblock \urlprefix\url{https://cohere.com/blog/introducing-embed-v3}.
\newblock \bibinfo{note}{Accessed: 2024-12-12}.

\bibitem{geron2022}
\bibinfo{author}{Géron, A.}
\newblock \emph{\bibinfo{title}{Hands-On Machine Learning with Scikit-Learn,
  Keras, and TensorFlow: Concepts, Tools, and Techniques to Build Intelligent
  Systems}} (\bibinfo{publisher}{O'Reilly Media}, \bibinfo{year}{2022}),
  \bibinfo{edition}{2nd} edn.

\bibitem{encord_contrastive_learning}
\bibinfo{author}{Blog, E.}
\newblock \bibinfo{title}{Guide to contrastive learning: Methods and
  applications}.
\newblock \emph{\bibinfo{journal}{Encord Blog}}  (\bibinfo{year}{2024}).
\newblock
  \urlprefix\url{https://encord.com/blog/guide-to-contrastive-learning/}.
\newblock \bibinfo{note}{Accessed: 2024-12-10}.

\bibitem{hastie2009}
\bibinfo{author}{Hastie, T.}, \bibinfo{author}{Tibshirani, R.} \&
  \bibinfo{author}{Friedman, J.}
\newblock \emph{\bibinfo{title}{The Elements of Statistical Learning: Data
  Mining, Inference, and Prediction}} (\bibinfo{publisher}{Springer},
  \bibinfo{year}{2009}).

\bibitem{breiman2001}
\bibinfo{author}{Breiman, L.}
\newblock \bibinfo{title}{Random forests}.
\newblock \emph{\bibinfo{journal}{Machine Learning}}
  \textbf{\bibinfo{volume}{45}}, \bibinfo{pages}{5--32} (\bibinfo{year}{2001}).

\bibitem{singh2024aya}
\bibinfo{author}{Singh, S.} \emph{et~al.}
\newblock \bibinfo{title}{Aya dataset: An open-access collection for
  multilingual instruction tuning} (\bibinfo{year}{2024}).
\newblock \eprint{2402.06619}.

\end{thebibliography}

\end{document}